\documentclass{article}
\usepackage{spconf,amsmath,graphicx,hyperref}
\usepackage{amsfonts}
\usepackage{booktabs}
\usepackage{tabularx}
\usepackage{multirow}

\title{Delving Deeper: Hierarchical Visual Perception for Robust Video-Text Retrieval}
\name{Zequn Xie$^{\star \dagger}$ \qquad  Boyun Zhang$^{\star \dagger}$ \qquad  Yuxiao Lin$^{\dagger}$ \qquad Tao Jin$^{\dagger}$}

\address{$^{\star}$ These authors contributed equally to this work. \\
         $^{\dagger}$ Zhejiang University \\
         The corresponding author is Tao Jin.
         }
\begin{document}
%
\maketitle
\begin{abstract}
Video-text retrieval (VTR) aims to locate relevant videos using natural language queries. Current methods, often based on pre-trained models like CLIP, are hindered by video's inherent redundancy and their reliance on coarse, final-layer features, limiting matching accuracy. To address this, we introduce the \textbf{HVP-Net (Hierarchical Visual Perception Network)}, a framework that mines richer video semantics by extracting and refining features from multiple intermediate layers of a vision encoder. Our approach progressively distills salient visual concepts from raw patch-tokens at different semantic levels, mitigating redundancy while preserving crucial details for alignment. This results in a more robust video representation, leading to new state-of-the-art performance on challenging benchmarks including MSRVTT, DiDeMo, and ActivityNet. Our work validates the effectiveness of exploiting hierarchical features for advancing video-text retrieval.Our codes are available at https://github.com/boyun-zhang/HVP-Net.
\end{abstract}

\section{Introduction}
\label{sec:intro}

Cross-modal retrieval aims to bridge the inherent heterogeneity gap between different data modalities by projecting them into a common latent space for similarity measurement \cite{han2024crossmodal, feng2014crossmodal}. This paradigm and advanced feature fusion techniques have been successfully applied across various domains, including text-based person retrieval \cite{xie2025dynamicuncertaintylearningnoisy, xie2025chat}, financial risk analysis \cite{ke2025early, ke2025stable}, fake news detection \cite{lu2025dammfnd, tong2025dapt}, and sequential recommendation \cite{lu2025dmmd4sr, cui2025multi}. In Video-Text Retrieval (VTR), this is often achieved by leveraging large pre-trained models like CLIP \cite{radford2021learningtransferablevisualmodels} to learn a joint video-text embedding space. However, a critical limitation persists: these approaches, even those employing more advanced multi-grained matching strategies \cite{wang2021t2vladgloballocalsequencealignment, Min2022HunYuantvrFT,zhang2022vldeformervisionlanguage}, predominantly rely on features from the \textit{final layer} of the vision encoder \cite{luo2021clip4clipempiricalstudyclip, yang2020treeaugmentedcrossmodalencodingcomplexquery, ma2022xclipendtoendmultigrainedcontrastive}. This practice leads to significant information redundancy by indiscriminately encoding all visual tokens. More importantly, it overlooks the rich, hierarchical semantics and dynamics encapsulated within the encoder's intermediate layers \cite{ lu2022understanding}, thereby constraining both representation efficiency and matching precision.

\begin{figure}[t!]

    \centering

    \includegraphics[width=\linewidth]{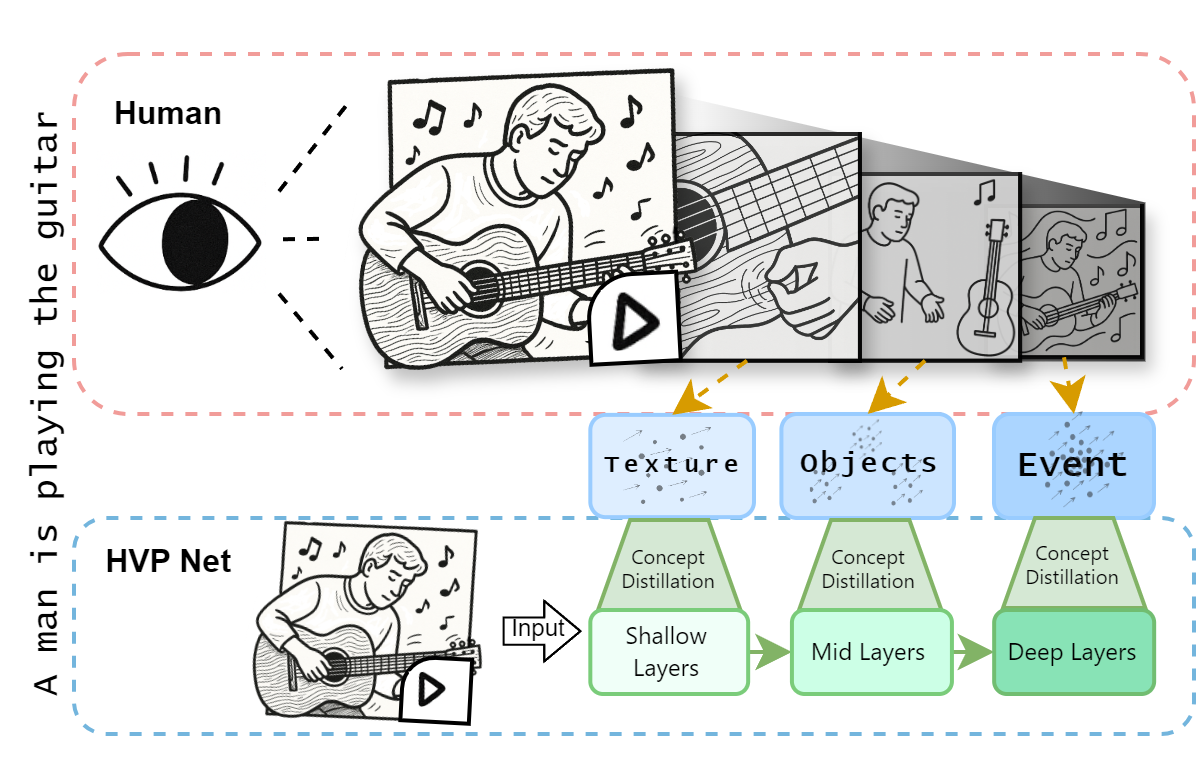} 

    \caption{Conceptual illustration of our HVP-Net, inspired by the human hierarchical perception process. Our model mimics this by processing visual data through shallow, mid, and deep layers to extract concepts ranging from low-level textures to high-level events.}

    \label{fig:explain}

\end{figure}

\begin{figure*}[t!]

    \centering

    \includegraphics[width=\linewidth]{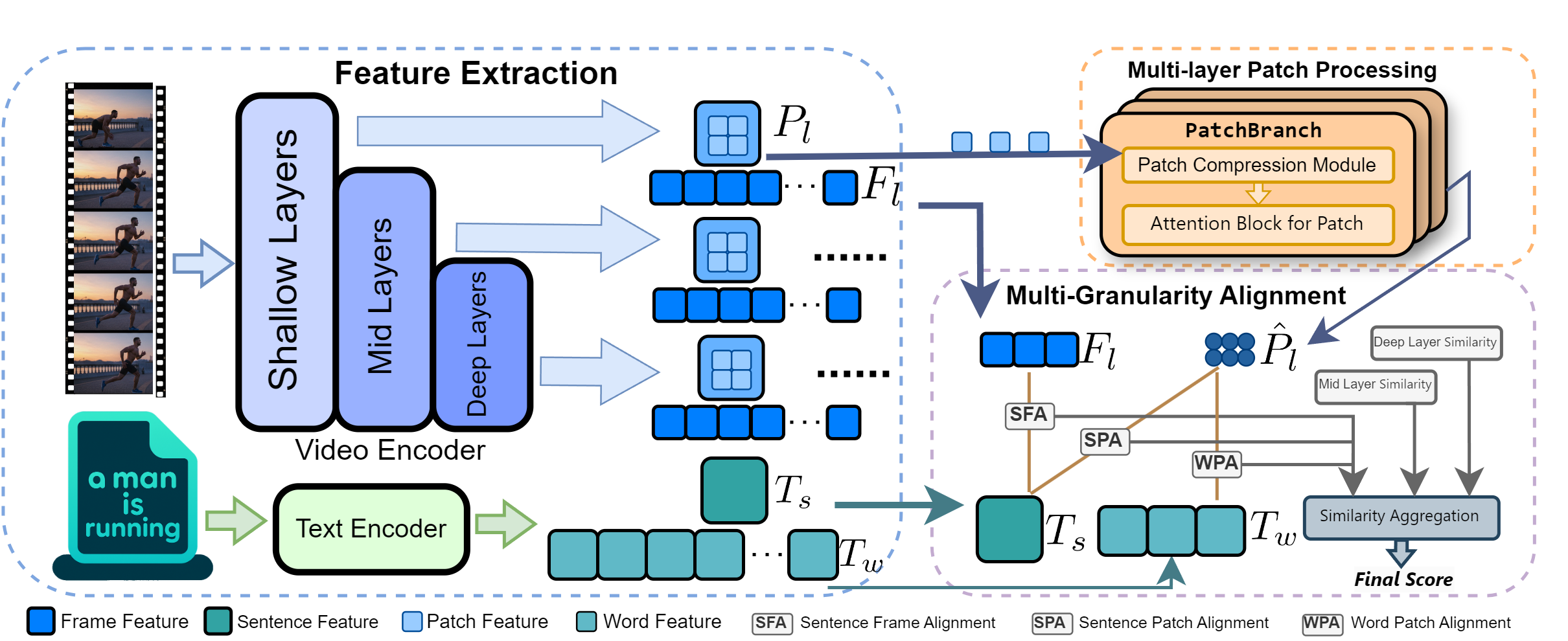} 

    \caption{The architecture of our HVP-Net. We first extract hierarchical frame ($F_l$) and patch ($P_l$) features from a video encoder. Then, our Multi-layer Patch Processing module refines the patch features into $\hat{P}_l$. Finally, we perform multi-granularity alignment between text and the multi-level video features ($F_l$, $\hat{P}_l$) and aggregate the layer-wise similarity scores for retrieval.}

    \label{fig:framework}

\end{figure*}
\section{Methodology}
\label{sec:methodology}
In this section, we detail our proposed \textbf{Hierarchical Visual Perception Network (HVP-Net)}. As illustrated in Figure~\ref{fig:framework}, our core idea is to construct a robust video representation by extracting and refining features from multiple intermediate layers of a pre-trained vision encoder. The process involves three main stages: hierarchical feature encoding, a novel patch processing module, and multi-granularity alignment optimized via contrastive learning.

\subsection{Hierarchical Feature Encoding}
\label{sec:encoding}
We use the pre-trained CLIP model \cite{radford2021learningtransferablevisualmodels} as our backbone. From its text encoder, we extract global sentence features $T_s \in \mathbb{R}^{D}$ and word-level features $T_w \in \mathbb{R}^{L_w \times D}$.

Distinct from prior work that relies on final-layer outputs, we extract features from a set of intermediate layers $\mathcal{L}$ of the Vision Transformer. For a video with $N$ frames, this provides hierarchical features for each layer $l \in \mathcal{L}$: frame features $F_l \in \mathbb{R}^{N \times D}$ (from [CLS] tokens) and patch features $P_l \in \mathbb{R}^{N \times M \times D}$ (from patch tokens, with $M$ patches per frame). This coarse-to-fine representation is then fed into our refinement module.

\subsection{Multi-layer Patch Processing (MPP)}
\label{sec:mpp}
To distill semantic concepts from redundant patch features, we introduce the \textbf{Multi-layer Patch Processing (MPP)} module, which iteratively refines patch features $P_l$ from each layer $l$. Each iteration involves a "compression-refinement" cycle:
\begin{enumerate}
    \item \textbf{Patch Compression Module} We first compute a saliency score for all $M$ patch tokens. A Density-Peak Clustering (DPC) algorithm then dynamically identifies $K$ cluster centers ($K \ll M$) from these tokens. Finally, tokens are merged into $K$ features via a saliency-weighted average based on cluster assignments.
    \item \textbf{Attention Block for Patch:} The $K$ compressed features act as queries in-a-cross-attention block, attending to the original $M$ tokens to enrich the compressed representations with fine-grained details.
\end{enumerate}
This iterative process outputs a purified patch feature set $\hat{P}_l \in \mathbb{R}^{N \times K \times D}$ for each layer, forming a distilled multi-level representation of visual entities.

\begin{figure*}[t!]
    \centering
    \includegraphics[width=\linewidth]{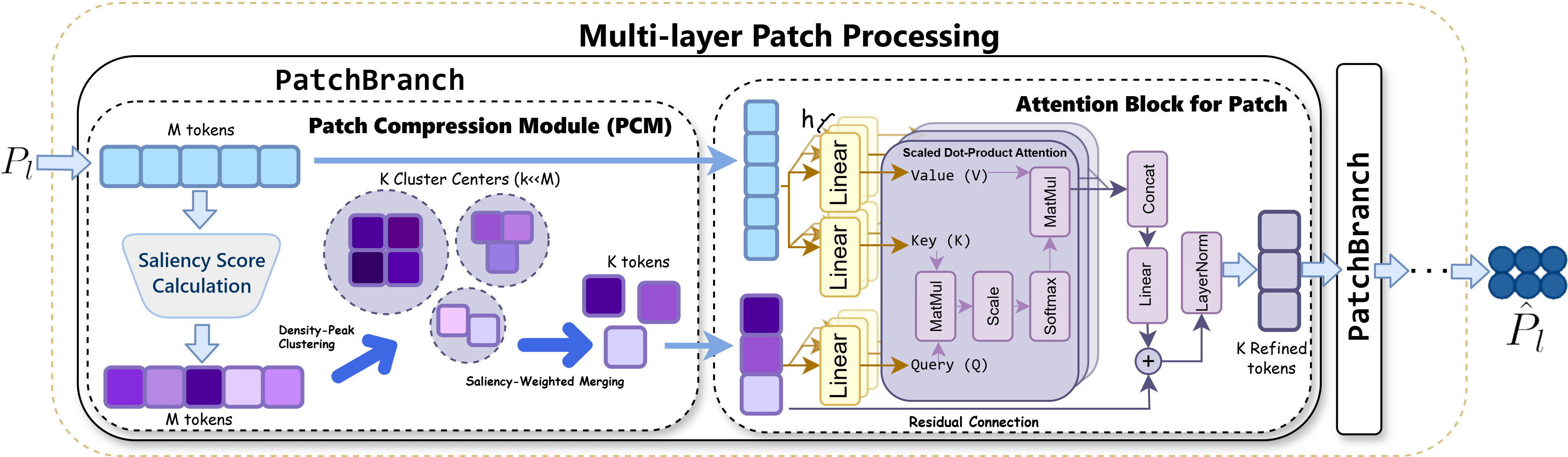} 
    \caption{The architecture of the Multi-layer Patch Processing (MPP) module, which iteratively refines input patch features $P_l$ to $\hat{P}_l$. In each processing block, a Patch Compression Module (PCM) first distills $M$ tokens into $K$ core concepts using saliency-guided clustering. Subsequently, these concepts are refined via a cross-attention mechanism that attends back to the original tokens.}
    \label{fig:mpp}
\end{figure*}

\subsection{Multi-granularity Alignment and Optimization}
\label{sec:alignment}
To ensure precise retrieval, we perform cross-modal alignment across three semantic granularities, spanning all $L$ extracted feature levels. For a given text-video pair $(i, j)$, we compute the following similarity scores for each layer $l$:

\begin{enumerate}
    \item \textbf{Sentence-Frame (SF) Alignment:} Similarity between the global sentence feature $T_{s_i}$ and the frame features $F_{l_j}$.
    \item \textbf{Sentence-Patch (SP) Alignment:} Similarity between $T_{s_i}$ and the refined patch concepts $\hat{P}_{l_j}$.
    \item \textbf{Word-Patch (WP) Alignment:} Similarity between word features $T_{w_i}$ and refined patch concepts $\hat{P}_{l_j}$.
\end{enumerate}

For patch-level and word-level features, the final similarity score is not derived from simple pooling. Instead, we first identify the most relevant patch for each word token (and vice versa) using max-pooling over the token-to-patch similarity matrix. Then, we employ learnable weights, predicted by small MLPs, to perform a weighted aggregation of these maximum similarity scores. This allows the model to dynamically emphasize the most salient word and patch contributions. For a given layer $l$, the word-to-patch similarity component is calculated as:
\begin{equation}
S_{\text{w} \to \text{p}}^{(l)}(i, j) = \sum_{k=1}^{N_w} W_{\text{MLP}}(w_{ik}) \cdot \max_{m=1}^{N_p} \text{sim}(w_{ik}, p_{jm})
\end{equation}
A symmetric patch-to-word similarity is also computed, and the final score is their average.

\textbf{Objective Function.} We optimize the model using a symmetric contrastive loss (InfoNCE) \cite{Oord2018RepresentationLW}. The total loss is the sum of bidirectional contrastive losses computed for each alignment granularity at each feature level:
\begin{equation}
\mathcal{L}_{\text{total}} = \sum_{l=1}^{L} (\mathcal{L}_{\text{SF}}^{(l)} + \mathcal{L}_{\text{SP}}^{(l)} + \mathcal{L}_{\text{WP}}^{(l)})
\label{eq:total_loss}
\end{equation}
where each component, e.g., $\mathcal{L}_{\text{SF}}^{(l)}$, is a standard InfoNCE loss calculated on the sentence-frame similarity matrix for layer $l$. This joint optimization across multiple levels and granularities ensures the learning of a consistent and robust cross-modal semantic space.

\textbf{Inference.} At inference time, the final similarity score between a query text and a candidate video is the sum of all similarity scores computed across all granularities and layers, maintaining consistency with the training objective.

\section{Experiments}
\label{sec:experiment}

We conduct extensive experiments to evaluate our proposed HVP-Net. We first introduce the experimental setup, then compare our method with state-of-the-art (SOTA) approaches, and finally provide in-depth analysis through ablation studies and qualitative examples.

\subsection{Experimental Setup}
\label{sec:setup}

\textbf{Datasets and Metrics.} Our evaluation is performed on three mainstream benchmarks: \textbf{MSR-VTT} \cite{xu2016msrvtt},  \textbf{DiDeMo} \cite{hendricks2017localizingmomentsvideonatural}, \textbf{ActivityNet Captions} \cite{Krishna2017dense}. For MSR-VTT, we follow the standard `1k-A` test split. We use standard retrieval metrics: \textbf{Recall at K (R@K, K=1, 5, 10)} (higher is better), \textbf{Median Rank (MdR)}, and \textbf{Mean Rank (MnR)} (lower is better), for both text-to-video (T2V) and video-to-text (V2T) tasks.

\textbf{Implementation Details.} Our model is built upon the CLIP ViT-B/32 backbone. We uniformly sample 12 frames per video at 224x224 resolution, with a maximum text length of 24 tokens. We use an initial learning rate of 1e-4 (a coefficient of 0.001 for the CLIP backbone) and a weight decay of 0.2. The model is trained for 5 epochs on MSR-VTT with a batch size of 32. The MPP module processes features extracted from three layers of the vision encoder: a shallow (1st), a middle (6th), and a deep (12th) Transformer layer.

\subsection{Comparison with State-of-the-Art Methods}
\label{sec:sota_comparison}

We compare HVP-Net against recent SOTA methods, including strong CLIP-based models like CLIP4Clip \cite{luo2021clip4clipempiricalstudyclip}, X-CLIP \cite{ma2022xclipendtoendmultigrainedcontrastive}, BiHSSP\cite{Liu_Huang_Xiong_Lv_2025}, MUSE\cite{tang2025musemambaefficientmultiscale}.

\begin{table}[h!]
\centering
\small
\renewcommand{\arraystretch}{0.85} 
\setlength{\tabcolsep}{3pt}
\caption{Performance comparison on the MSR-VTT 1k-A test set. R@K is in percentage (\%). Best results are in \textbf{bold}.}
\label{tab:sota_msrvtt}
\begin{tabularx}{\linewidth}{@{} X ccccc @{}}
\toprule
\textbf{Method} & \textbf{R@1 $\uparrow$} & \textbf{R@5 $\uparrow$} & \textbf{R@10 $\uparrow$} & \textbf{MdR $\downarrow$} & \textbf{MnR $\downarrow$} \\
\midrule
\multicolumn{6}{l}{\textit{Text-to-Video (T2V)}} \\
CLIP4Clip \cite{luo2021clip4clipempiricalstudyclip} & 44.5 & 71.4 & 81.6 & 2.0 & 15.3 \\
X-CLIP \cite{ma2022xclipendtoendmultigrainedcontrastive} & 49.3 & 75.8 & 84.8 & 2.0 & 12.2 \\
BiHSSP\cite{Liu_Huang_Xiong_Lv_2025}& 48.1& 74.0& 84.1& 2.0& 12.1\\
MUSE\cite{tang2025musemambaefficientmultiscale}& 50.9& 76.7& 85.6& 1.0& 10.9\\
\textbf{HVP-Net (ours)} & \textbf{56.7} & \textbf{83.9} & \textbf{90.6} & \textbf{1.0} & \textbf{5.3} \\
\midrule
\multicolumn{6}{l}{\textit{Video-to-Text (V2T)}} \\
CLIP4Clip \cite{luo2021clip4clipempiricalstudyclip} &42.7&70.9&80.6&2.0&11.6 \\
X-CLIP \cite{ma2022xclipendtoendmultigrainedcontrastive} & 48.9 & 76.8 & 84.5 & 2.0 & 8.1 \\
BiHSSP\cite{Liu_Huang_Xiong_Lv_2025}& 48.0& 74.1& 83.5& 2.0& 9.0\\
MUSE\cite{tang2025musemambaefficientmultiscale}& 49.7& 77.8& 86.5& 2.0& 7.4\\
\textbf{HVP-Net  (ours)} & \textbf{50.5} & \textbf{80.5} & \textbf{91.2} & \textbf{2.0} & \textbf{4.4} \\
\bottomrule
\end{tabularx}
\end{table}

As shown in Table~\ref{tab:sota_msrvtt}, HVP-Net significantly surpasses all competitors on MSR-VTT. For text-to-video retrieval, our method achieves an R@1 of 56.7\%, outperforming the strongest baseline MUSE by a substantial 5.8\% margin. This leading performance is consistent across all metrics for both text-to-video and video-to-text tasks. Table~\ref{tab:sota_dide_act} further confirms our model’s superiority and generalization, as it establishes new state-of-the-art results on both DiDeMo and ActivityNet, outperforming prior works across all reported recall metrics.

\begin{table}[h!]
\centering
\renewcommand{\arraystretch}{0.85}
\setlength{\tabcolsep}{1pt} 
\caption{Performance comparison with state-of-the-art methods on DiDeMo and ActivityNet for text-to-video retrieval.}
\label{tab:sota_dide_act}
\begin{tabular}{@{}l|ccc|ccc@{}}
\toprule
\multirow{2}{*}{\textbf{Method}} & \multicolumn{3}{c|}{\textbf{DiDeMo}} & \multicolumn{3}{c}{\textbf{ActivityNet}} \\
\cmidrule(lr){2-4} \cmidrule(lr){5-7}
 & R@1↑ & R@5↑ & R@10↑ & R@1↑ & R@5↑ & R@10↑ \\
\midrule
Clip4Clip \cite{luo2021clip4clipempiricalstudyclip}     & -    & -    & -    & 40.5 & 72.4 & 83.6 \\
CLIP-VIP \cite{xue2022clipvip}         & 48.6 & 77.1 & 84.4 & -    & -    & -    \\
HBI \cite{jin2023video}               & -    & -    & -    & 42.2 & 73.0 & 84.6 \\
DiffusionRet \cite{jin2023diffusionret} & 46.7 & 74.7 & 82.7 & -    & -    & -    \\
\midrule
\textbf{HVP-Net (Ours)} & \textbf{57.1} & \textbf{83.1} & \textbf{87.0} & \textbf{43.5} & \textbf{78.3} & \textbf{88.2} \\
\bottomrule
\end{tabular}
\end{table}

\subsection{Ablation Studies and Analysis}
\label{sec:ablation_analysis}
We conduct comprehensive ablation studies on the MSR-VTT dataset to analyze the key components of HVP-Net. All results for the T2V task are reported in Table~\ref{tab:ablation_all}.

\textbf{Effectiveness of Core Components.}
We first validate our two main contributions: the multi-layer strategy and the Multi-layer Patch Processing module. As shown in Table~\ref{tab:ablation_all} (left), a baseline using only final-layer features achieves an R@1 of 54.2\%. The full HVP-Net model boosts performance to 56.7\%. Notably, simply aggregating raw intermediate features without our proposed MPP module (+ Multi-layer (w/o MPP)) leads to a dramatic performance collapse (22.1\%), confirming that our purification process is essential for effectively fusing multi-layer information.

\textbf{Impact of Layer Selection.}
We analyze different layer selection strategies in Table~\ref{tab:ablation_all} (middle). While using contiguous Last 3 layers (55.6\%) improves over the Last 1 layer baseline (54.2\%), our default strategy of sampling from diverse semantic stages ({1, 6, 12}) performs best (56.7\%). This supports our hypothesis that fusing low-level details with high-level semantics is beneficial. Using more layers ({1, 3, 6, 9, 12}) degrades performance (51.9\%), indicating our three-stage sampling is an efficient and effective choice.

\textbf{Impact of Alignment Losses.}
Finally, ablating each loss component (Table~\ref{tab:ablation_all}, right) confirms their necessity. Removing the sentence-frame ($\mathcal{L}_{SF}$), sentence-patch ($\mathcal{L}_{SP}$), or word-patch ($\mathcal{L}_{WP}$) alignment losses leads to performance drops of \textbf{8.5}, \textbf{3.5}, and a substantial \textbf{23.4} points, respectively. This demonstrates that all three supervision levels are crucial, with the fine-grained word-patch alignment being the most impactful component for robust cross-modal understanding.

\begin{table}[h!]
\centering
\caption{Ablation studies on core components, layer selection strategy, and loss functions (T2V R@1, \%). $\Delta$ denotes the performance drop from the full model.}
\label{tab:ablation_all}
\resizebox{\columnwidth}{!}{%
\begin{tabular}{@{}lc|lc|lc@{}}
\toprule
\textbf{Component Ablation} & \textbf{R@1} & \textbf{Layer Selection} & \textbf{R@1} & \textbf{Loss Ablation} & \textbf{R@1 ($\Delta$)} \\
\cmidrule(r){1-2} \cmidrule(lr){3-4} \cmidrule(l){5-6}
Baseline (Last 1 layer) & 54.2 & Last 1 layer & 54.2 & Full Model & 56.7 (-) \\
+ Multi-layer (w/o MPP) & 22.1 & Last 3 layers & 55.6 & w/o $\mathcal{L}_{SF}$ & 48.2 (-8.5) \\
\textbf{+ MPP (Full Model)} & \textbf{56.7} & 5 layers {1,3,6,9,12} & 51.9 & w/o $\mathcal{L}_{SP}$ & 53.2 (-3.5) \\
& & \textbf{Default {1,6,12}} & \textbf{56.7} & w/o $\mathcal{L}_{WP}$ & 33.3 (-23.4) \\
\bottomrule
\end{tabular}%
}
\end{table}

\section{Conclusion}

In this paper, we addressed the critical challenge of information redundancy in text-video retrieval, a limitation of conventional methods that rely solely on final-layer features. We proposed the \textbf{HVP-Net}, a novel framework featuring a \textbf{Multi-layer Patch Processing (MPP) Module}. This module effectively distills salient concepts by refining features from the intermediate layers of a vision encoder, creating a more robust cross-modal embedding space. Our approach establishes new state-of-the-art performance on multiple benchmarks, including a notable R@1 of 56.7\% on MSR-VTT. Our work validates that mining the internal hierarchical features of pre-trained models is a potent strategy for cross-modal alignment. Future work could explore adaptive layer-selection strategies and specific mechanisms to improve video-to-text retrieval performance.




\bibliographystyle{IEEEbib}
\bibliography{strings,refs}

@misc{radford2021learningtransferablevisualmodels,
      title={Learning Transferable Visual Models From Natural Language Supervision}, 
      author={Alec Radford and Jong Wook Kim and Chris Hallacy and Aditya Ramesh and Gabriel Goh and Sandhini Agarwal and Girish Sastry and Amanda Askell and Pamela Mishkin and Jack Clark and Gretchen Krueger and Ilya Sutskever},
      year={2021},
      eprint={2103.00020},
      archivePrefix={arXiv},
      primaryClass={cs.CV},
      url={https://arxiv.org/abs/2103.00020}, 
}

@misc{xie2025dynamicuncertaintylearningnoisy,
      title={Dynamic Uncertainty Learning with Noisy Correspondence for Text-Based Person Search}, 
      author={Zequn Xie and Haoming Ji and Chengxuan Li and Lingwei Meng},
      year={2025},
      eprint={2505.06566},
      archivePrefix={arXiv},
      primaryClass={cs.CV},
      url={https://arxiv.org/abs/2505.06566}, 
}

@inproceedings{lu2022understanding,
 title={Understanding the dynamics of dnns using graph modularity},
 author={Lu, Yao and Yang, Wen and Zhang, Yunzhe and Chen, Zuohui and Chen, Jinyin and Xuan, Qi and Wang, Zhen and Yang, Xiaoniu},
 booktitle={European Conference on Computer Vision},
 pages={225--242},
 year={2022},
 organization={Springer}
}

@article{ke2025early,
  title={Early warning of cryptocurrency reversal risks via multi-source data},
  author={Ke, Zong and Cao, Yuqing and Chen, Zhenrui and Yin, Yuchen and He, Shouchao and Cheng, Yu},
  journal={Finance Research Letters},
  pages={107890},
  year={2025},
  publisher={Elsevier}
}

@inproceedings{lu2025dammfnd,
  title={DAMMFND: Domain-Aware Multimodal Multi-view Fake News Detection},
  author={Lu, Weihai and Tong, Yu and Ye, Zhiqiu},
  booktitle={Proceedings of the AAAI Conference on Artificial Intelligence},
  volume={39},
  number={1},
  pages={559--567},
  year={2025}
}

@inproceedings{lu2025dmmd4sr,
  title={DMMD4SR: Diffusion Model-based Multi-level Multimodal Denoising for Sequential Recommendation},
  author={Lu, Weihai and Yin, Li},
  booktitle={Proceedings of the 33rd ACM International Conference on Multimedia},
  pages={6363--6372},
  year={2025}
}

@inproceedings{tong2025dapt,
  title={DAPT: Domain-Aware Prompt-Tuning for Multimodal Fake News Detection},
  author={Tong, Yu and Lu, Weihai and Cui, Xiaoxi and Mao, Yifan and Zhao, Zhejun},
  booktitle={Proceedings of the 33rd ACM International Conference on Multimedia},
  pages={7902--7911},
  year={2025}
}

@inproceedings{cui2025multi,
  title={Multi-modal multi-behavior sequential recommendation with conditional diffusion-based feature denoising},
  author={Cui, Xiaoxi and Lu, Weihai and Tong, Yu and Li, Yiheng and Zhao, Zhejun},
  booktitle={Proceedings of the 48th International ACM SIGIR Conference on Research and Development in Information Retrieval},
  pages={1593--1602},
  year={2025}
}

@article{ke2025stable,
  title={A stable technical feature with GRU-CNN-GA fusion},
  author={Ke, Zong and Shen, Jiaqing and Zhao, Xuanyi and Fu, Xinghao and Wang, Yang and Li, Zichao and Liu, Lingjie and Mu, Huailing},
  journal={Applied Soft Computing},
  pages={114302},
  year={2025},
  publisher={Elsevier}
}

@inproceedings{xie2025chat,
  title={Chat-Driven Text Generation and Interaction for Person Retrieval},
  author={Xie, Zequn and Wang, Chuxin and Wang, Yeqiang and Cai, Sihang and Wang, Shulei and Jin, Tao},
  booktitle={Proceedings of the 2025 Conference on Empirical Methods in Natural Language Processing},
  pages={5259--5270},
  year={2025}
}

@misc{luo2021clip4clipempiricalstudyclip,
      title={CLIP4Clip: An Empirical Study of CLIP for End to End Video Clip Retrieval}, 
      author={Huaishao Luo and Lei Ji and Ming Zhong and Yang Chen and Wen Lei and Nan Duan and Tianrui Li},
      year={2021},
      eprint={2104.08860},
      archivePrefix={arXiv},
      primaryClass={cs.CV},
      url={https://arxiv.org/abs/2104.08860}, 
}

@misc{yang2020treeaugmentedcrossmodalencodingcomplexquery,
      title={Tree-Augmented Cross-Modal Encoding for Complex-Query Video Retrieval}, 
      author={Xun Yang and Jianfeng Dong and Yixin Cao and Xun Wang and Meng Wang and Tat-Seng Chua},
      year={2020},
      eprint={2007.02503},
      archivePrefix={arXiv},
      primaryClass={cs.CV},
      url={https://arxiv.org/abs/2007.02503}, 
}

@misc{wang2021t2vladgloballocalsequencealignment,
      title={T2VLAD: Global-Local Sequence Alignment for Text-Video Retrieval}, 
      author={Xiaohan Wang and Linchao Zhu and Yi Yang},
      year={2021},
      eprint={2104.10054},
      archivePrefix={arXiv},
      primaryClass={cs.CV},
      url={https://arxiv.org/abs/2104.10054}, 
}

@misc{ma2022xclipendtoendmultigrainedcontrastive,
      title={X-CLIP: End-to-End Multi-grained Contrastive Learning for Video-Text Retrieval}, 
      author={Yiwei Ma and Guohai Xu and Xiaoshuai Sun and Ming Yan and Ji Zhang and Rongrong Ji},
      year={2022},
      eprint={2207.07285},
      archivePrefix={arXiv},
      primaryClass={cs.CV},
      url={https://arxiv.org/abs/2207.07285}, 
}

@article{Oord2018RepresentationLW,
  title={Representation Learning with Contrastive Predictive Coding},
  author={A{\"a}ron van den Oord and Yazhe Li and Oriol Vinyals},
  journal={ArXiv},
  year={2018},
  volume={abs/1807.03748},
  url={https://api.semanticscholar.org/CorpusID:49670925}
}

@INPROCEEDINGS{Krishna2017dense,
  author={Krishna, Ranjay and Hata, Kenji and Ren, Frederic and Fei-Fei, Li and Niebles, Juan Carlos},
  booktitle={2017 IEEE International Conference on Computer Vision (ICCV)}, 
  title={Dense-Captioning Events in Videos}, 
  year={2017},
  volume={},
  number={},
  pages={706-715},
  doi={10.1109/ICCV.2017.83}}

@INPROCEEDINGS{xu2016msrvtt,
  author={Xu, Jun and Mei, Tao and Yao, Ting and Rui, Yong},
  booktitle={2016 IEEE Conference on Computer Vision and Pattern Recognition (CVPR)}, 
  title={MSR-VTT: A Large Video Description Dataset for Bridging Video and Language}, 
  year={2016},
  volume={},
  number={},
  pages={5288-5296},
  doi={10.1109/CVPR.2016.571}}

@inproceedings{Min2022HunYuantvrFT,
  title={HunYuan\_tvr for Text-Video Retrivial},
  author={Shaobo Min and Weijie Kong and Rong-Cheng Tu and Dihong Gong and Chengfei Cai and Wenzhe Zhao and Chenyang Liu and Sixiao Zheng and Hongfa Wang and Zhifeng Li and Wei Liu},
  year={2022},
  url={https://api.semanticscholar.org/CorpusID:248006223}
}

@misc{hendricks2017localizingmomentsvideonatural,
      title={Localizing Moments in Video with Natural Language}, 
      author={Lisa Anne Hendricks and Oliver Wang and Eli Shechtman and Josef Sivic and Trevor Darrell and Bryan Russell},
      year={2017},
      eprint={1708.01641},
      archivePrefix={arXiv},
      primaryClass={cs.CV},
      url={https://arxiv.org/abs/1708.01641}, 
}

@article{Liu_Huang_Xiong_Lv_2025, title={Learning Dynamic Similarity by Bidirectional Hierarchical Sliding Semantic Probe for Efficient Text Video Retrieval}, volume={39}, url={https://ojs.aaai.org/index.php/AAAI/article/view/32604}, DOI={10.1609/aaai.v39i6.32604}, abstractNote={Text-video retrieval is a foundation task in multi-modal research which aims to align texts and videos in the embedding space. The key challenge is to learn the similarity between videos and texts. A conventional approach involves directly aligning video-text pairs using cosine similarity. However, due to the disparity in the information conveyed by videos and texts, i.e., a single video can be described from multiple perspectives, the retrieval accuracy is suboptimal. An alternative approach employs cross-modal interaction to enable videos to dynamically acquire distinct features from various texts, thus facilitating similarity calculations. Nevertheless, this solution incurs a computational complexity of O(n^2) during retrieval. To this end, this paper proposes a novel method called Bidirectional Hierarchical Sliding Semantic Probe (BiHSSP), which calculates dynamic similarity between videos and texts with O(n) complexity during retrieval. We introduce a hierarchical semantic probe module that learns semantic probes at different scales for both video and text features. Semantic probe involves a sliding calculation of the cross-correlation between semantic probes at different scales and embeddings from another modality, allowing for dynamic similarity computation between video and text descriptions from various perspectives. Specifically, for text descriptions from different angles, we calculate the similarity at different locations within the video features and vice versa. This approach preserves the complete information of the video while addressing the issue of unequal information between video and text without requiring cross-modal interaction. Additionally, our method can function as a plug-and-play module across various methods, thereby enhancing the corresponding performance. Experimental results demonstrate that our BiHSSP significantly outperforms the baseline.}, number={6}, journal={Proceedings of the AAAI Conference on Artificial Intelligence}, author={Liu, Yang and Huang, Shudong and Xiong, Deng and Lv, Jiancheng}, year={2025}, month={Apr.}, pages={5667-5675} }

@misc{tang2025musemambaefficientmultiscale,
      title={MUSE: Mamba is Efficient Multi-scale Learner for Text-video Retrieval}, 
      author={Haoran Tang and Meng Cao and Jinfa Huang and Ruyang Liu and Peng Jin and Ge Li and Xiaodan Liang},
      year={2025},
      eprint={2408.10575},
      archivePrefix={arXiv},
      primaryClass={cs.CV},
      url={https://arxiv.org/abs/2408.10575}, 
}

@inproceedings{xue2022clipvip,
  title={Clip-vip: Adapting pre-trained image-text model to video-language alignment},
  author={Xue, Hong and Sun, Yuchong and Liu, Bei and Fu, Jianlong and Song, Rui and Li, Houqiang and Luo, Jiebo},
  booktitle={The Eleventh International Conference on Learning Representations},
  year={2022}
}

@inproceedings{jin2023video,
  title={Video-text as game players: Hierarchical banzhaf interaction for cross-modal representation learning},
  author={Jin, Peng and Huang, Jin and Xiong, Peng and Tian, Shizhe and Liu, Chang and Ji, Xiang and Yuan, Li and Chen, Jian},
  booktitle={Proceedings of the IEEE/CVF Conference on Computer Vision and Pattern Recognition},
  pages={2472--2482},
  year={2023}
}

@inproceedings{jin2023diffusionret,
  title={Diffusionret: Generative text-video retrieval with diffusion model},
  author={Jin, Peng and Li, Hao and Cheng, Zekai and Li, Ke and Ji, Xiang and Liu, Chang and Yuan, Li and Chen, Jian},
  booktitle={Proceedings of the IEEE/CVF international conference on computer vision},
  pages={2470--2481},
  year={2023}
}

@article{han2024crossmodal,
  author={Han, Zhichao and Azman, Azreen Bin and Mustaffa, Mas Rina Binti and Khalid, Fatimah Binti},
  journal={IEEE Access}, 
  title={Cross-Modal Retrieval: A Review of Methodologies, Datasets, and Future Perspectives}, 
  year={2024},
  volume={12},
  pages={115716-115741},
  doi={10.1109/ACCESS.2024.3444817}
}

@inproceedings{feng2014crossmodal,
  title={Cross-modal retrieval with correspondence autoencoder},
  author={Feng, Fang and Wang, Xiaofei and Li, Rushi},
  booktitle={Proceedings of the 22nd ACM international conference on Multimedia},
  pages={7--16},
  year={2014}
}

@article{zhang2022vldeformervisionlanguage,
  title={VLDeformer: Vision--language decomposed transformer for fast cross-modal retrieval},
  author={Zhang, Lisai and Wu, Hong and Chen, Qing and Deng, Yuan and Siebert, Jochen and Li, Zili and Han, Yuhan and Kong, Dong and Cao, Zhe},
  journal={Knowledge-Based Systems},
  volume={252},
  pages={109316},
  year={2022},
  publisher={Elsevier}
}

\end{document}